\documentclass[runningheads]{llncs}
\usepackage[T1]{fontenc}
\usepackage{graphicx}

\usepackage{colortbl}
\usepackage{amsmath, amssymb}
\usepackage{multirow}
\usepackage{xcolor}

\usepackage{color}
\usepackage{url}

\usepackage{booktabs}

\usepackage{orcidlink}
\newcommand\orcid[1]{\textsuperscript{\orcidlink{#1}}}

\begin{document}
\title{Principled Context Engineering for RAG: Statistical Guarantees via Conformal Prediction}
\titlerunning{Principled Context Engineering for RAG}

\author{Debashish Chakraborty \inst{1}\orcid{0009-0000-8656-9406} \and
 Eugene Yang \inst{1}\orcid{0000-0002-0051-1535} \and
 Daniel Khashabi \inst{2}\orcid{0009-0009-7664-2230} \and
 Dawn Lawrie \inst{1}\orcid{0000-0001-7347-7086} \and
 Kevin Duh\inst{1}\orcid{0000-0001-8107-4383}}

\authorrunning{Chakraborty et al.}

\institute{
HLTCOE, Johns Hopkins University, Baltimore, Maryland, USA \\
\email{\{dchakra6,eugene.yang,lawrie,kduh1\}@jhu.edu}
\and
Johns Hopkins University, Baltimore, Maryland, USA \\
\email{danielk@cs.jhu.edu}\\
Correspondence: \email{dchakra6@jhu.edu}
}

\maketitle


\begin{abstract}

Retrieval-Augmented Generation (RAG) enhances factual grounding in large language models (LLMs) 
by incorporating retrieved evidence, but LLM accuracy declines when long or noisy contexts exceed 
the model's effective attention span. Existing pre-generation filters 
rely on heuristics or uncalibrated LLM confidence scores, offering no statistical control 
over retained evidence.  
We evaluate and demonstrate \emph{context engineering through conformal prediction}, 
a coverage-controlled filtering framework that removes irrelevant content while preserving 
recall of supporting evidence. Using both embedding- and LLM-based scoring functions, 
we test this approach on the NeuCLIR and RAGTIME collections. Conformal filtering 
consistently meets its target coverage, ensuring that a specified fraction of 
relevant snippets are retained, and reduces retained context by 2--3$\times$ relative 
to unfiltered retrieval. On NeuCLIR, downstream factual accuracy measured by ARGUE~F1 
improves under strict filtering and remains stable at moderate coverage, indicating that 
most discarded material is redundant or irrelevant. 
These results demonstrate that conformal prediction enables reliable, coverage-controlled context 
reduction in RAG, offering a model-agnostic and principled approach to context engineering.


\end{abstract}


\section{Introduction}\label{sec:Introduction}

Retrieval-Augmented Generation (RAG) grounds large language models (LLMs) in retrieved evidence, 
reducing hallucinations compared to standalone models~\cite{lewis2021rag,shuster-etal-2021-retrieval-augmentation}. 
Despite rapid progress, RAG systems remain brittle, with retrieval noise and prompt saturation degrading reliability
~\cite{Magesh2024HallucinationFreeAT,Asgari2024.09.12.24313556}. 
The \textit{lost-in-the-middle} effect~\cite{liu-etal-2024-lost} shows that LLMs attend poorly to mid-prompt evidence, 
limiting effective use of long-context capacity to 10--20\%~\cite{Hsieh2024RULERWT,hong2025context}.
Context has therefore been reframed as a finite \emph{attention budget}, motivating high-signal, compact inputs 
for reliable generation~\cite{anthropic2025contextengineering}.
Retrieval noise compounds this issue. 
Most vector databases rank text by cosine similarity between dense embeddings, 
yet such similarity scores are typically uncalibrated and may be weakly correlated with true relevance~\cite{Steck2024IsCO}. 
Irrelevant or marginally related passages frequently enter the prompt, diluting useful evidence and inflating token costs. 
Benchmarks such as RAGTruth~\cite{niu-etal-2024-ragtruth} and CRAG \cite{10.5555/3737916.3738251} show that such errors harm factual accuracy. 

We address these challenges by introducing \textbf{context engineering through conformal prediction} 
as a principled mechanism for pre-generation filtering in RAG. Conformal prediction provides 
finite-sample coverage guarantees, ensuring that a specified proportion of relevant snippets 
are retained while irrelevant material is filtered out without additional model training
\cite{vovk2005algorithmic,Angelopoulos2021AGI}. Unlike prior RAG calibration methods that 
operate post-generation, our approach applies conformal prediction immediately after retrieval, 
offering formal control over context composition and context size.

We evaluate the framework on NeuCLIR~\cite{Lawrie2025OverviewOT} and RAGTIME~\cite{trecragtime2025}. 
Across both, conformal filtering achieves target coverage while reducing context size by 2--3$\times$. 
On NeuCLIR, answer quality measured by ARGUE~F1~\cite{mayfield2024argue} improves under strict filtering 
and remains stable at moderate levels, indicating that most removed content contributes little to downstream generation. 
Together, these results show that conformal prediction enables reliable, coverage-controlled context reduction, 
establishing it as a lightweight, model-agnostic foundation for principled context engineering in RAG. Our work makes three contributions:
\begin{enumerate}
    \item We introduce a framework for \textbf{context engineering} in RAG, 
    applying conformal prediction after retrieval to guarantee coverage of relevant evidence.
    \item We empirically demonstrate that \textbf{conformal filtering} achieves target coverage while 
    reducing context size by 2--3$\times$ across NeuCLIR and RAGTIME, maintaining factual accuracy 
    under strict filtering.
    \item We show that this approach is \textbf{model-agnostic},
    needs no retraining, and works with both embedding- and LLM-based scoring functions.
\end{enumerate} 


\section{Related Work}
\label{sec:relatedwork}

Prior research on mitigating retrieval noise in RAG systems can be grouped into 
three categories: heuristic filtering, LLM-based re-ranking, and conformal calibration. 
In this section, we review limitations of each class in turn.

\paragraph{Heuristic Filtering.}
Most production RAG pipelines rely on simple heuristics such as top-$k$
retrieval or fixed similarity thresholds. Frameworks like LlamaIndex and vector
databases such as Weaviate rank chunks by cosine distance
\cite{llamaindex_embeddings_guide,weaviateVectorIndexing}. 
While efficient, these methods
may exhibit different effectiveness across topics \cite{Steck2024IsCO}.
Empirical studies find that irrelevant or marginally related snippets frequently 
pass such filters, diluting evidence and amplifying long-context degradation
\cite{niu-etal-2024-ragtruth,Kang2024CRAGCG}.

\paragraph{LLM-Based Filtering.}
Recent work has explored using the generator itself to assess snippet quality.
LLatrieval prompts LLMs to judge retrieval sufficiency
\cite{li-etal-2024-llatrieval}, MiniCheck employs LLMs for claim-level
verification \cite{tang-etal-2024-minicheck}, and other pipelines decouple
evidence selection from generation
\cite{slobodkin-etal-2024-attribute,Sohn2024RationaleGuidedRA}. 
LLM confidence values are not probabilistic posteriors and are frequently
miscalibrated -- though monotonically correlated with relevance -- they form coarse, 
prompt-sensitive scales that lack statistical calibration
\cite{guo2017calibration,miao2024canllmsuncertainty,zhang2025evaluatinguncertainty,Lovering2024LanguageMP}.

\paragraph{Conformal Prediction for RAG.}

Conformal prediction (CP) provides finite-sample, distribution-free guarantees
on coverage without assuming a calibrated posterior
\cite{vovk2005algorithmic,Angelopoulos2021AGI}. Recent studies extend CP to
RAG systems: \textbf{Conformal-RAG} improves group-conditional coverage for
claim verification \cite{Feng2025ResponseQA}, while \textbf{C-RAG} bounds
the risk of factual error during generation relative to standalone LLMs
\cite{Kang2024CRAGCG}. 
Closest to our setting, CONFLARE \cite{Rouzrokh2024CONFLARECL} calibrates similarity 
cutoffs to control retrieval uncertainty 
at the retrieval stage, while TRAQ \cite{li-etal-2024-traq} provides an end-to-end 
correctness guarantee over answer sets 
in retrieval-augmented Question Answering. 
Our contribution differs in scope: we apply split conformal prediction directly to snippet retention 
immediately after retrieval and evaluate its coverage efficiency trade-off (and downstream 
nugget-based factuality where available) under topic-disjoint calibration/test splits.
By applying CP immediately after retrieval, we prevent noisy or redundant content from entering 
the generator, enabling on-the-fly filtering 
that constrains context length while ensuring coverage.


\section{Methodology}

\begin{figure}[t]
\centering
\includegraphics[width=\linewidth]{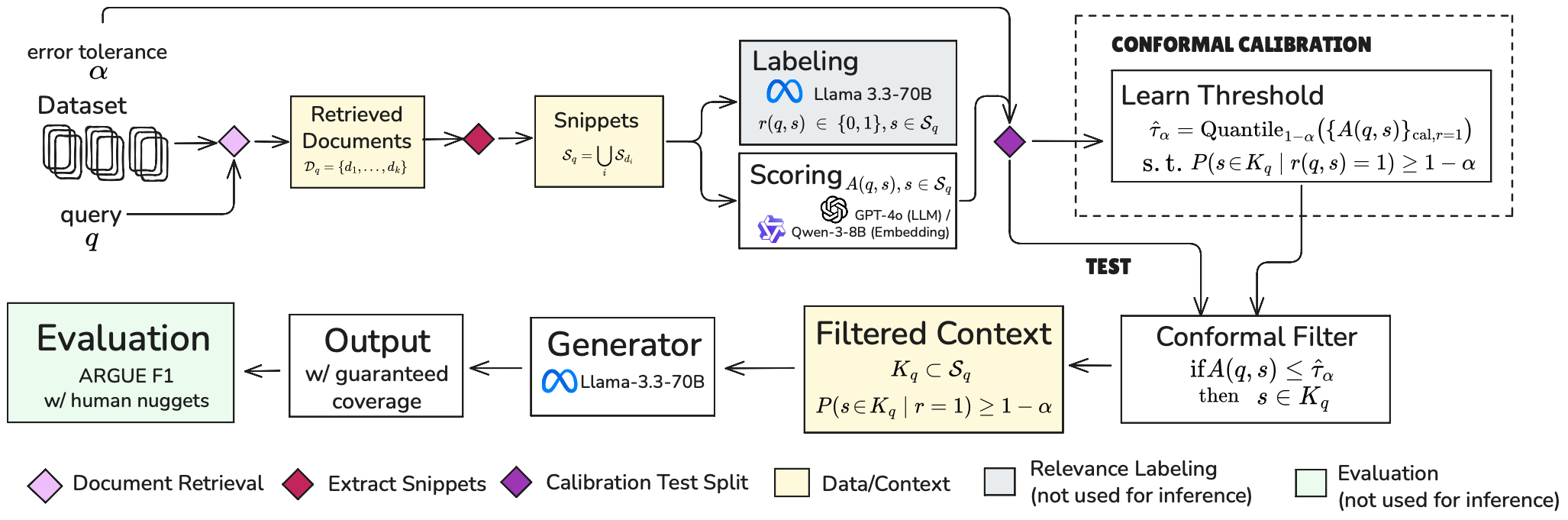}
\caption{\textbf{Conformal context filtering workflow.}
Query $q$ retrieves documents $\mathcal{D}_q$, segmented into snippets $\mathcal{S}_q$.
Each snippet is scored $A(q,s)$, calibrated to threshold $\hat{\tau}_\alpha$,
filtered to $K_q = \{s:A(q,s)\le\hat{\tau}_\alpha\}$, and passed to generation.
ARGUE F1 evaluates generated answers against human nuggets.}
\label{fig:workflow}
\end{figure}

\subsection{Problem Formulation}

Given a query $q$, a retriever returns a set of documents 
$\mathcal{D}_q = \{d_1,\dots,d_k\}$ that may contain both relevant and 
irrelevant content, motivating snippet-level filtering. 
Each document is segmented into 500-character windows overlapping by 
100 characters total (50 on each side), preserving sentence boundaries 
following empirical chunking analysis~\cite{Brdland2025ANH}. 
Let $\mathcal{S}_q$ denote all retrieved snippets and 
$r(q,s) \in \{0,1\}$ indicate whether snippet $s$ supports answering $q$.
We aim to construct a filtered subset $K_q \subseteq \mathcal{S}_q$ that retains 
relevant snippets under a user-specified \emph{miscoverage rate} 
$\alpha\!\in\!(0,1)$. 
Formally, the selection rule must achieve marginal coverage
$
P(s \in K_q \mid r(q,s)=1) \ge 1-\alpha,
$
ensuring that a labeled-relevant snippet $(q,s)$ drawn exchangeably with calibration is retained with probability at least $1-\alpha$.
A lower $\alpha$ provides stronger coverage guarantees at the cost of 
including more context, 
while higher $\alpha$ allows more aggressive filtering.
Figure~\ref{fig:workflow} summarizes the workflow.

\subsection{Split Conformal Prediction for Context Filtering}

We apply split conformal prediction
\cite{Angelopoulos2021AGI,vovk2005algorithmic,Lei2016DistributionFreePI} to obtain finite-sample
marginal coverage guarantees.
The method requires a \textbf{scoring function} $A(q,s)$ assigning
nonconformity scores (lower = more relevant), a labeled
\textbf{calibration set} $\mathcal{D}_{\text{cal}}$, and a disjoint
\textbf{test set} $\mathcal{D}_{\text{test}}$ from the same distribution.

Given miscoverage $\alpha$, the empirical $(1-\alpha)$-quantile of the positive
calibration scores defines the filtering threshold:
\[
\hat{\tau}_\alpha =
\text{Quantile}_{1-\alpha} \big(\{A(q,s):(q,s) \in \mathcal{D}_{\text{cal}},r(q,s) = 1\}\big),
\]
or equivalently, the $\lceil(n{+}1)(1{-}\alpha)\rceil$-th smallest score among
$n$ positive examples.  
At test time, a snippet $s$ is retained iff $A(q,s) \le \hat{\tau}_\alpha$.
Under the exchangeability assumption between calibration and test
splits, this guarantees $P(s \in K_q\mid r(q,s) = 1) \ge 1{-}\alpha$.

\subsection{Experimental Setup}

We now describe the experimental setup used to evaluate this framework.
We evaluate on \textbf{NeuCLIR}~\cite{Lawrie2025OverviewOT} and 
\textbf{RAGTIME}~\cite{trecragtime2025}, using disjoint query topics for 
calibration and test to preserve exchangeability 
(NeuCLIR: 1{,}440/740 snippets; RAGTIME: 1{,}710/560).

\paragraph{Scoring functions.}
We test two paradigms:
\begin{enumerate}
    \item \textbf{Conformal-Embedding} using Qwen3-Embedding-8B \cite{Zhang2025Qwen3EA}, $A_{\text{emb}}(q,s) = 1 - \cos(\mathrm{emb}(q), \mathrm{emb}(s))$; and
    \item \textbf{Conformal-LLM} using GPT-4o \cite{Hurst2024GPT4oSC} prompted to rate snippet relevance on $[0,1]$, $A_{\text{LLM}}(q,s) = 1 - \text{rating}$
\end{enumerate}

\paragraph{Relevance labeling.}
Calibration and test relevance labels $r(q,s)$ are generated by
Llama 3.3-70B-Instruct \cite{Dubey2024TheL3} using a rubric-style prompt, 
similar to \cite{tang-etal-2024-minicheck}, asking whether each snippet
supports the query.  The model outputs a binary decision parsed into
$r(q,s) \in \{0,1\}$.
Calibration labels define $\hat{\tau}_\alpha$; test labels are used only for
empirical coverage evaluation.
A 10\% subsample was manually reviewed to verify consistency between human judgments and model labels.

\paragraph{Labeling as an annotation function.}
We treat the labeler as an \emph{annotation function} that provides consistent
binary relevance labels for conformal calibration and empirical coverage
measurement. Human nuggets in NeuCLIR are used only for downstream evaluation
(ARGUE~F1) and are never used to set conformal thresholds. As with any automatically 
generated annotation, guarantees are conditional on
label consistency across calibration and test topics. 
The exact prompt templates and output format used for LLM relevance scoring and labeling are released in our repository.\footnote{\url{https://github.com/hltcoe/conformal-context-engineering}}

\paragraph{Generation and evaluation.}
Filtered snippets $K_q$ are concatenated and provided to the same
Llama 3.3-70B generator for answer production to maintain consistency between
labeling and generation.
We report:
(1) empirical coverage,
(2) removal rate, and
(3) downstream factual quality on NeuCLIR via ARGUE F1 \cite{mayfield2024argue} with the AutoARGUE implementation~\cite{Walden2025AutoARGUELR}.
Since the nugget annotation of the RAGTIME collection, which is used for TREC RAGTIME Track in 2025, was not available at the time of conducting our experiments, it is therefore evaluated only for coverage and removal behavior.


\section{Results and Discussion}

\begin{figure}[t]
\centering
\includegraphics[width=0.9\linewidth]{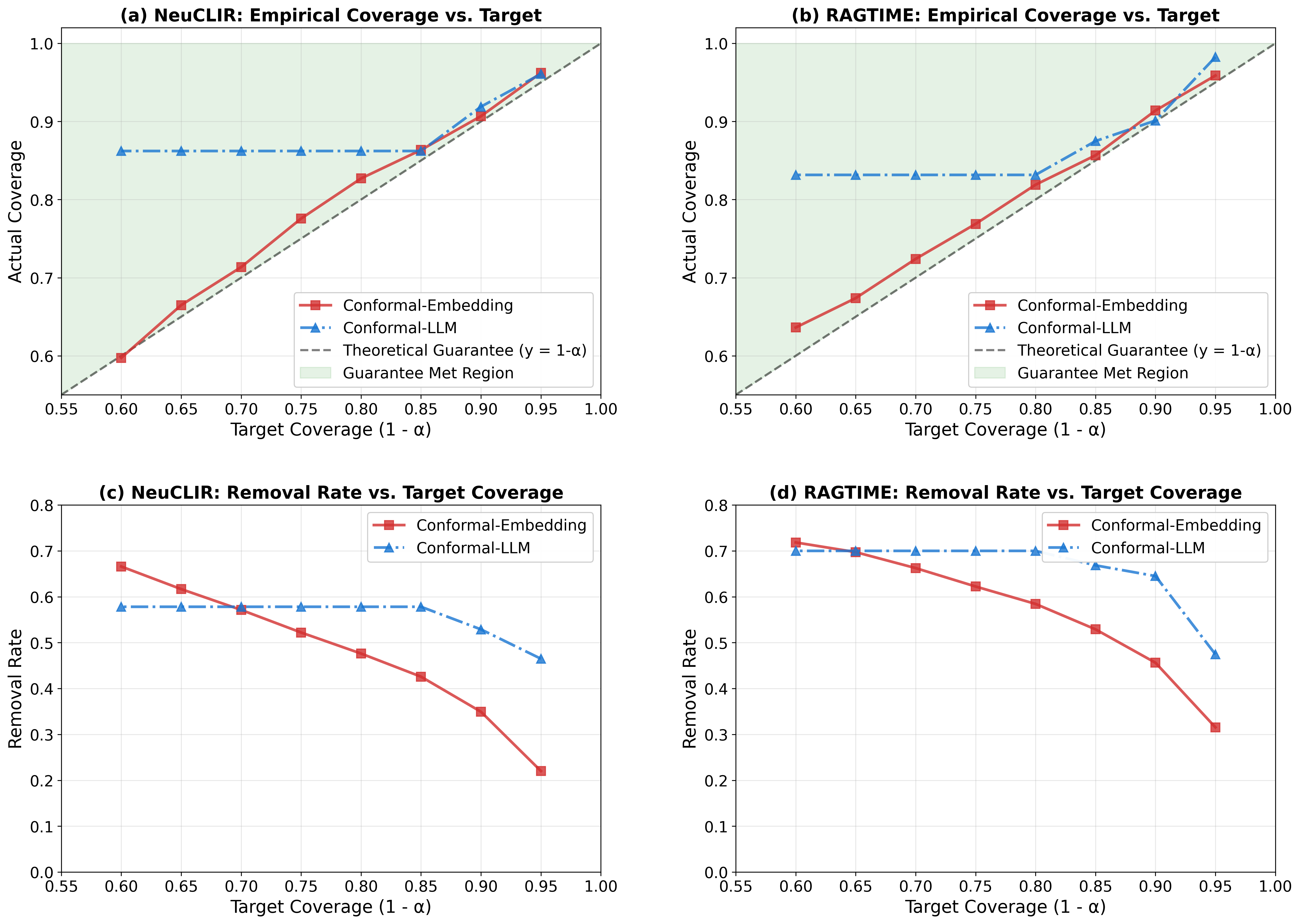}
\caption{\textbf{Coverage guarantees and context reduction across NeuCLIR and RAGTIME.}
(a–b)~Empirical coverage vs.~target $(1-\alpha)$ (dashed: theoretical guarantee, shaded: valid region). 
Both methods satisfy conformal guarantees; Conformal-Embedding follows the target line more closely. 
(c–d)~Removal rate vs.~$(1-\alpha)$, illustrating the expected monotonic trade-off between tighter coverage and
stronger filtering. Conformal-LLM removes more context overall but in quantized steps.
}
\label{fig:coverage_efficiency}
\end{figure}

\paragraph{Coverage Guarantees.}
Figure~\ref{fig:coverage_efficiency}(a–b) shows empirical coverage against the
target $(1-\alpha)$ for \textbf{NeuCLIR} and \textbf{RAGTIME}. 
Across all $\alpha$ values (0.05–0.40), both Conformal-Embedding and Conformal-LLM
meet or slightly exceed theoretical coverage guarantees, confirming the
finite-sample validity of split conformal prediction. 
Conformal-LLM exhibits mild over-coverage (flat segments near 0.85–0.87) due to 
discretized rating bins, yielding coarser control over coverage levels. 
By contrast, Conformal-Embedding tracks the target line more smoothly, providing
finer adaptation to coverage. 

\paragraph{Heuristic pruning baselines.}
As a quantitative reference, per-query top-$k$ snippet pruning yields uncontrolled coverage:
on NeuCLIR, $k{=}30$ achieves 30\% context reduction but only 76\% empirical coverage, and
$k{=}25$ achieves 39\% reduction but only 68\% coverage (w.r.t.\ $r(q,s)$), well below the 90–95\%
coverage levels targeted by conformal filtering.

\paragraph{Heuristic threshold baselines.}
Fixed cosine thresholds provide no coverage control and vary widely across queries:
on NeuCLIR, $\theta{=}0.50$ yields 76\%$\pm$20\% (mean $\pm$ std across queries) coverage at 35\% context reduction, while
$\theta{=}0.60$ drops coverage to 43\%$\pm$15\% at 69\% reduction.
In contrast, conformal filtering targets a user-specified coverage level (e.g., 90\%) and tracks it closely under exchangeability.

\paragraph{Context Reduction Efficiency.}

Figures~\ref{fig:coverage_efficiency}(c--d) present removal rate as a function
of target coverage. Removal decreases monotonically as $(1-\alpha)$ increases,
illustrating the expected trade-off between tighter guarantees and stronger
filtering.  
At strict coverage ($\alpha \le 0.20$), Conformal-Embedding removes
25--55\% of retrieved snippets while maintaining full coverage, offering stable
and interpretable control of context size.  
Conformal-LLM removes 46--70\% of content but exhibits discrete jumps in
removal rate due to its quantized confidence ratings.  
This consistent monotonic behavior across both datasets demonstrates that
conformal filtering provides a reliable and tunable mechanism for managing
retrieval depth.

\paragraph{Downstream Answer Quality.}

We assess the impact of filtering on factual generation quality on NeuCLIR using
ARGUE~F1~\cite{mayfield2024argue} with AutoARGUE~\cite{Walden2025AutoARGUELR}.
The unfiltered baseline achieves 0.69~F1.   
Both filters improve ARGUE F1 at strict coverage (0.05–0.10) and
remain indistinguishable from the baseline at $\alpha{=}0.20$.
Together with the coverage results, these findings show that conformal filtering
effectively denoises retrieved context, removing redundant or weakly relevant
snippets without harming factual accuracy.  
\textbf{Table~\ref{tab:argue_neuclir_compact}} presents ARGUE~F1 and context reduction
side by side for each $\alpha$, highlighting the balance between factual
retention and filtering efficiency.

\begin{table}[t]
\centering
\caption{\textbf{NeuCLIR factual quality (ARGUE F1) and context reduction (ConRed\%).}
Bold marks the best value per column. 
$\dagger$ indicates significant improvement over the unfiltered baseline ($p{<}0.05$, paired bootstrap resampling). 
The unfiltered baseline achieves 0.69~F1 with 0\% reduction. 
}
\label{tab:argue_neuclir_compact}
\setlength{\tabcolsep}{3pt}
\begin{tabular}{@{}lccc@{}}
\toprule
\textbf{Method} &
\begin{tabular}[c]{@{}c@{}}$\alpha{=}0.05$\\(F1 / ConRed\%)\end{tabular} &
\begin{tabular}[c]{@{}c@{}}$\alpha{=}0.10$\\(F1 / ConRed\%)\end{tabular} &
\begin{tabular}[c]{@{}c@{}}$\alpha{=}0.20$\\(F1 / ConRed\%)\end{tabular} \\
\midrule
Conformal-Embedding & \textbf{0.720$^{\dagger}$} / 22.2 & \textbf{0.700}$^{\dagger}$  / 35.0 & 0.680 / 52.8 \\
Conformal-LLM & 0.710$^{\dagger}$ / 46.5 & \textbf{0.700}$^{\dagger}$  / 58.0 & 0.680 / 57.8 \\
\bottomrule
\end{tabular}
\end{table}

\paragraph{Discussion.}

The results highlight three observations:
\textbf{(1)} 
Split conformal prediction reliably enforces coverage guarantees
across datasets and scoring models, turning pre-generation filtering from a
heuristic into a statistically grounded process.  
\textbf{(2)} 
Conformal-Embedding provides smooth, predictable control at high coverage targets
(80--95\% marginal coverage of labeled-relevant snippets), whereas Conformal-LLM achieves stronger pruning 
but in coarse steps due to score quantization.
\textbf{(3)}
On NeuCLIR, ARGUE~F1 improves at strict coverage 
($\alpha\!\in\!\{0.05,0.10\}$) and remains statistically indistinguishable from the baseline at 
$\alpha{=}0.20$, indicating that more than half of retrieved snippets can be pruned
without loss in factual quality. 
Although absolute gains are modest, the stability itself is informative: once
retrieval noise is reduced, the generator likely operates near its effective
attention limit.  
This finding reframes conformal prediction 
as a practical tool for \emph{context engineering}, enabling 
robust, coverage-aware filtering before generation, laying the foundation for 
adaptive recalibration under domain or topic shifts.


\section{Conclusion}
We presented a statistical framework for \emph{context engineering} in RAG 
based on split conformal prediction. 
Across NeuCLIR and RAGTIME, both embedding- and LLM-based conformal filters achieved guaranteed coverage 
while reducing context size by up to threefold. 
On NeuCLIR, downstream factual quality (ARGUE~F1) improved under strict filtering and remained stable 
at moderate coverage, showing that redundant content can be safely pruned without loss of accuracy. 
These findings demonstrate that conformal prediction enables reliable, coverage-controlled context reduction 
and provides a lightweight, model-agnostic foundation for scalable RAG. 
Future work will explore adaptive recalibration across topics and domains to relax the exchangeability 
assumption and extend statistical guarantees under distribution shift.


{\begin{credits}

\subsubsection{\discintname}
The authors have no competing interests to declare that are relevant to the
content of this article.

\end{credits}}

\bibliographystyle{splncs04}
\bibliography{mybibliography}

\end{document}